%% file: main_with_supplement.tex
\documentclass[letterpaper]{article}
\usepackage[preprint]{aaai2027}
\usepackage[hyphens]{url}
\usepackage{graphicx}
\usepackage{natbib}
\usepackage{caption}
\usepackage{booktabs}
\usepackage{amsmath}
\usepackage{amssymb}
\usepackage{amsthm}
\usepackage{algorithm}
\usepackage{algorithmic}
\usepackage{xcolor}
\usepackage{multirow}
\usepackage{dblfloatfix}
\usepackage{docmute}

\copyrighttext{Preprint. Under review.}

\title{Does More Retrieved Evidence Help Visual Retrieval-Augmented Generation with Diffusion Language Models?}
\author{
    Jiankun Wang\textsuperscript{\rm 1},
    Yisen Gao\textsuperscript{\rm 2},
    Ziwei Zhang\textsuperscript{\rm 1},\\
    Xingcheng Fu\textsuperscript{\rm 3},
    Jiaxin Bai\textsuperscript{\rm 4},
    Chen Gao\textsuperscript{\rm 5}
}
\affiliations{
    \textsuperscript{\rm 1}School of Computer Science and Engineering, Beihang University\\
    \textsuperscript{\rm 2}Department of Computer Science and Engineering, The Hong Kong University of Science and Technology\\
    \textsuperscript{\rm 3}School of Computer Science and Engineering, Guangxi Normal University\\
    \textsuperscript{\rm 4}Department of Computer Science, Hong Kong Baptist University\\
    \textsuperscript{\rm 5}School of Computing, National University of Singapore
}

\newcommand{\method}{ECF}
\newcommand{\gain}{G}
\newcommand{\blank}{\mathrm{blank}}
\newcommand{\calH}{\mathcal{H}}
\newcommand{\KL}{\mathrm{KL}}
\newtheorem{theorem}{Theorem}
\newtheorem{proposition}[theorem]{Proposition}
\newtheorem{corollary}[theorem]{Corollary}
\theoremstyle{definition}
\newtheorem{definition}{Definition}

\theoremstyle{plain}

\begin{document}

\let\combinedendabstract\endabstract
\renewcommand{\endabstract}{%
  \par\noindent Code is publicly available at \url{https://github.com/wjkuser/ECF}.%
  \combinedendabstract}

\input{main.tex}

\clearpage
\begingroup
\setcounter{secnumdepth}{0}

\let\maketitle\relax
\def\bibliography#1{}
\newcommand{\combinedappendixlayout}{}
\newcommand{\beginlargerpooltable}{\begin{table*}[!b]}

\input{technical_supplement.tex}

\endgroup

\end{document}

%% file: main.tex
\maketitle

\begin{abstract}
Visual retrieval-augmented generation (RAG) commonly expands the retrieved evidence set to improve answer-page coverage, implicitly assuming that all available evidence should be passed to the generator.
We show that this assumption does not hold for diffusion language models (DLMs): retrieving more pages increases answer-page recall, whereas unconditionally passing all retrieved pages to the generator often reduces answer accuracy, primarily because of semantic conflict.
A latent-source analysis explains this mismatch through source-coherence loss in parallel denoising, where position-wise proposals can combine incompatible visual sources into unsupported answers.
We further find that such interference is already visible in the first-step answer-block distribution, making it possible to assess evidence before decoding.
To preserve retrieval coverage while limiting harmful visual exposure, we propose the Entropy-Based Candidate Filter (\method{}), a training-free evidence-admission framework.
To reduce irrelevant content within individual candidates, \method{} constructs multi-granularity evidence units; to identify beneficial additional evidence, it uses blank-controlled block confidence and retrieval rank to determine whether and which candidate should enter the final context.
Across three multimodal DLMs and five visual QA benchmarks, \method{} improves answer accuracy by 2.62 percentage points on average over the strongest fixed top-$k$ input and, with LLaDA2.0-Uni, by 2.37 percentage points on average over the best competing training-free result for each dataset.
These results show that broader retrieval benefits visual DLM-RAG through selective evidence admission rather than unconditional evidence expansion.
\end{abstract}

\section{Introduction}

Most modern large language models generate text autoregressively and have achieved strong performance across question answering, reasoning and tool-use tasks~\cite{xiao2026sing,ma2025general}.
More recently, diffusion language models (DLMs) have attracted increasing attention for their bidirectional conditioning and parallel decoding capabilities \citep{austin2021d3pm,hoogeboom2021argmax,sahoo2024simple,nie2025large,gao2026unifying}.
Recent studies \cite{inclusionai2026llada2uni,you2025lladav,ye2025dreamvl} have further extended DLMs to multimodal generation, enabling visual inputs to guide the denoising process.

However, many knowledge-intensive visual questions cannot be answered reliably from model parameters alone, creating a need for external evidence.
Retrieval-augmented generation (RAG) addresses this need by conditioning generation on retrieved content \citep{izacard2021fid,shi2024replug,asai2024selfrag}, and has been extended to visual tasks \citep{yu2024visrag,li2025regionrag}.
Visual RAG is challenging because answer-bearing evidence is often localized within a visually dense page, while each retrieved page introduces a complete visual source containing additional structured content.
To compensate for imperfect retrieval, existing visual RAG systems, which are predominantly built around autoregressive LLMs, often provide the generator with multiple top-ranked pages to improve answer-page coverage \citep{yu2024visrag,luo2026uoves}.
Recent work has improved the quality of this context through finer-grained region retrieval and generation-aware evidence selection \citep{li2025regionrag,luo2026uoves}.
Yet their implications for multimodal DLMs remain unclear: when several retrieved images are provided, they jointly condition all unresolved answer positions during parallel denoising.
Whether an individually relevant or useful candidate remains beneficial after entering this shared masked state is not well understood.
This motivates our central question: \emph{does providing more retrieved evidence actually improve visual RAG with diffusion language models?}

\begin{figure}[t]
\centering
\includegraphics[width=\columnwidth]{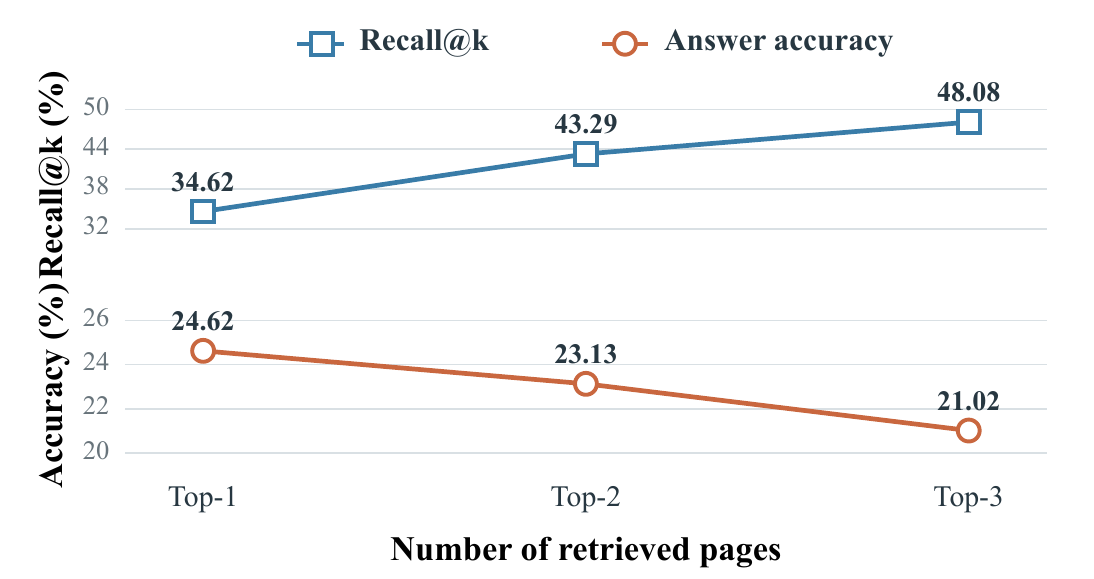}
\caption{Answer-page Recall@$k$ and answer accuracy under fixed top-$k$ input.}
\label{fig:intro-retrieval-accuracy}
\end{figure}

\begin{figure}[t]
\centering
\includegraphics[width=\columnwidth]{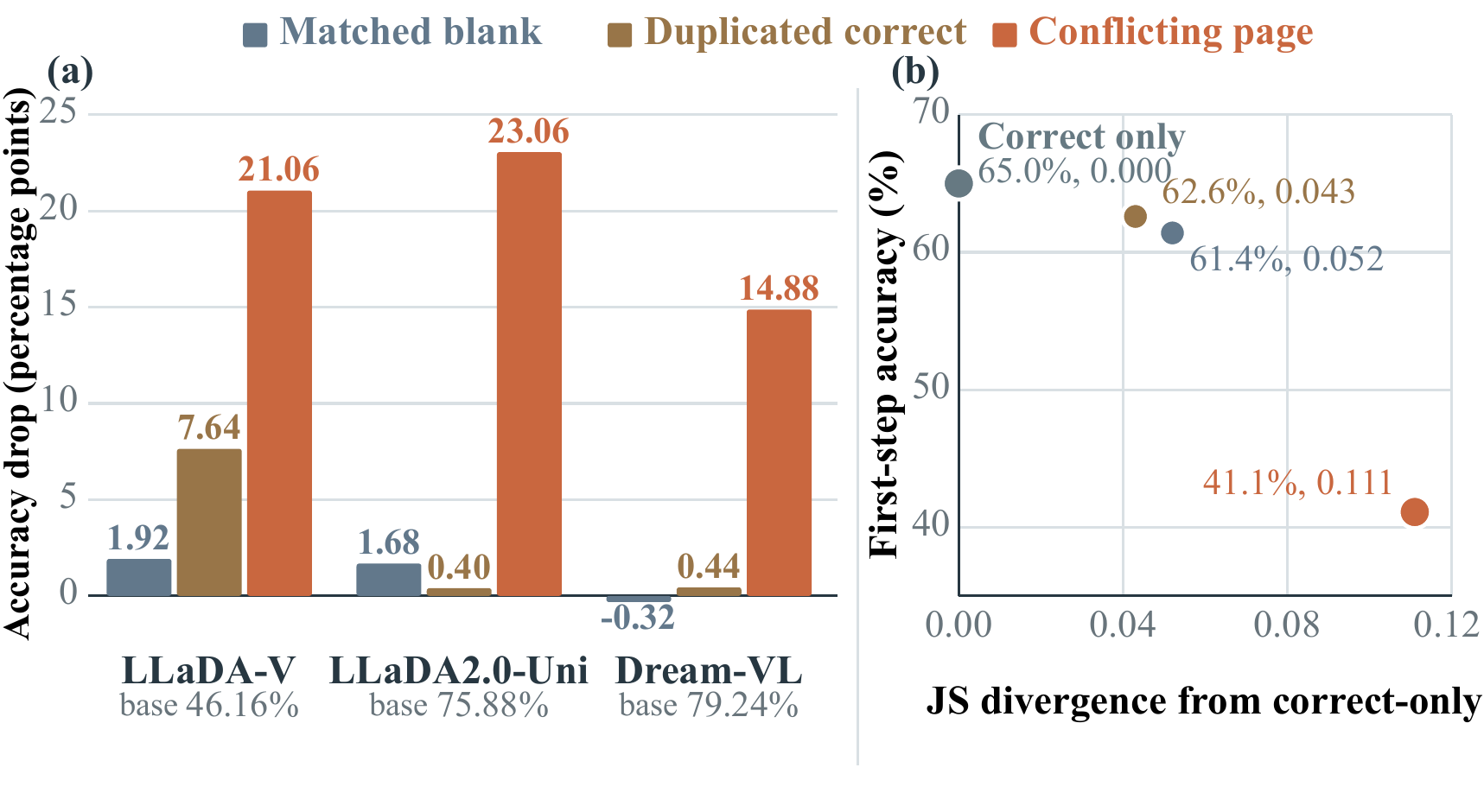}
\caption{Effects of additional visual evidence on ChartQA. (a) Final-answer accuracy drops relative to correct-only input under controlled second-image interventions across three backends; correct-only accuracies are shown below the backend names. (b) First-step answer-block accuracy and Jensen--Shannon divergence from the correct-only distribution on natural retrieval cases.}
\label{fig:intro-semantic-conflict}
\end{figure}

Our experiments show that more retrieved evidence does not necessarily help.
To examine how fixed input size affects retrieval coverage and generation quality, we evaluate fixed top-$k$ input with $k\in\{1,2,3\}$ using the same VisRAG-ranked candidate pools, prompts, decoding budgets, and answer scorer across three multimodal DLMs and five visual QA benchmarks.
Figure~\ref{fig:intro-retrieval-accuracy} reports macro-average answer-page Recall@$k$ across the five benchmarks and macro-average answer accuracy across all 15 model--dataset settings: Recall increases with $k$, whereas accuracy decreases as more pages are provided.
Figure~\ref{fig:intro-semantic-conflict}(a) further shows that conflicting pages cause the largest accuracy drop across all three backends, indicating that semantic conflict, rather than the mere presence of multiple images, drives the degradation.
Thus, expanding the candidate pool improves answer-page availability, but indiscriminately admitting additional candidates can make generation less reliable.

To explain this mismatch, Section~\ref{sec:source-theory} develops a restricted latent-source analysis of the first unresolved denoising state.
When correct and conflicting pages support different coherent answers, multiple unresolved positions share a latent source choice, whereas a position-wise factorized proposal does not preserve this coupling.
The analysis shows that this dependence loss assigns probability to cross-source answers supported by neither page, with the unsupported mass increasing as conflict spans more answer positions.
This identifies source-coherence loss as a mechanism behind the observed availability--accuracy mismatch.
Crucially, this failure is already visible before final decoding.
On naturally retrieved ChartQA cases, conflicting pages cause the largest drop in first-step answer-block accuracy and the largest Jensen--Shannon divergence from the correct-only distribution in Figure~\ref{fig:intro-semantic-conflict}(b); the same condition also produces the largest final-answer degradation in Figure~\ref{fig:intro-semantic-conflict}(a).
This early-to-final correspondence motivates the first-step block distribution as a generator-native signal for candidate assessment.

We therefore propose the \textbf{E}ntropy-Based \textbf{C}andidate \textbf{F}ilter (\textbf{\method{}}), a training-free framework that controls visual evidence exposure along two complementary axes.
For dense documents, multi-granularity evidence construction combines full pages with layout-derived regions to make localized evidence retrievable.
For generator-side admission, \method{} evaluates each additional candidate in the context in which it would actually be used.
Given the top-ranked evidence $c_1$ and a candidate $c_i$, it compares the target DLM's first-step block entropy under $\{c_1,c_i\}$ and $\{c_1,\blank(c_i)\}$, yielding a conditional confidence gain that controls for the structural effect of an additional visual input.
Under the local conditions in Proposition~\ref{prop:entropy-risk}, a positive gain corresponds to lower competing-source risk.
Rank-prior selection then uses these gains to decide whether to expand beyond $c_1$ and, if so, which candidate to admit.
Across three multimodal DLMs and five visual QA benchmarks, \method{} improves over the strongest fixed top-$k$ input by 2.62 percentage points on average.
With LLaDA2.0-Uni, it further outperforms the strongest training-free alternative by 2.37 percentage points on average.

Our contributions are:
\begin{itemize}
    \item We identify an availability--accuracy mismatch in visual DLM-RAG: broader retrieval improves answer-page coverage but can reduce answer accuracy. We trace this failure to source-coherence loss in parallel denoising.
    \item We introduce \method{}, which assesses additional evidence through a geometry-matched counterfactual in the target DLM's first-step answer block. Together with multi-granularity evidence construction, it decouples candidate-pool size from final-context size.
    \item Across five visual QA benchmarks and three multimodal DLMs, we show that selective evidence admission provides more reliable gains than fixed top-$k$ input and remains stable as the candidate pool expands.
\end{itemize}

\section{Related Work}

\paragraph{Visual RAG under imperfect evidence.}
Visual RAG retrieves external evidence directly from document images, preserving layout, graphics, and textual content that may be lost in text-only pipelines \citep{yu2024visrag,li2025regionrag}.
Existing work has improved visual evidence quality along two directions: stronger page-level retrieval and fusion, and finer-grained retrieval of query-relevant regions.
RegionRAG develops the latter direction by learning query--region alignment, and shows that irrelevant content can arise both within an individual page and across multiple retrieved pages \citep{li2025regionrag}.
Related robustness studies likewise find that irrelevant or poorly placed context can degrade generation \citep{yoran2024retrobust,yu2024chainnote,liu2024lostmiddle}.
These studies establish evidence granularity and candidate redundancy as important sources of noise, but do not characterize how multiple visual sources interact once they jointly condition diffusion generation.

\paragraph{Diffusion language models and DLM-RAG.}
Discrete and masked diffusion language models generate through iterative denoising \citep{austin2021d3pm,hoogeboom2021argmax,sahoo2024simple,nie2025large,gao2026unifying} rather than autoregressive factorization \citep{gao2025controllable,xiong2024autoregressive}, and recent work extends this paradigm to multimodal generation \citep{inclusionai2026llada2uni,you2025lladav,ye2025dreamvl}.
Because multiple positions can be updated from a shared masked state, their behavior depends on denoising order and on dependencies not represented by isolated position-wise marginals \citep{ni2026flexibility,kang2025parallelbench,zhang2026generation}.
For DLM-RAG, SPREAD uses query relevance to guide reveal order, while SARDI updates retrieval from tentative predictions \citep{yu2026spread,junger2026sardi}.
These methods study how available evidence is used or refreshed during denoising, rather than which visual candidates should jointly enter the initial masked context or how incompatible sources interact within it.

\paragraph{Adaptive evidence selection.}
Adaptive RAG systems decide when to retrieve and which evidence to retain using retrieval scores, generation uncertainty, self-reflection, learned verifiers, or surrogate multimodal models \citep{asai2024selfrag,jiang2023flare,luo2026uoves}.
UOVES provides the most direct generation-aware formulation for visual evidence selection: it defines evidence utility as information gain on the model's output distribution and derives a latent helpfulness objective that can be estimated by a lightweight surrogate in a training-free manner \citep{luo2026uoves}.
This formulation moves beyond semantic relevance, but operationally ranks candidates by their individual helpfulness and retains a prescribed top-$K$ subset.
It does not directly address context-dependent admission---whether an additional candidate remains beneficial when combined with already available visual evidence, or whether context expansion should be rejected before DLM denoising.
This work addresses this gap by studying evidence admission for DLMs whose unresolved answer positions share a multi-source visual context.

\section{Why More Retrieved Evidence Can Hurt Visual DLM-RAG}

\subsection{Preliminaries: Masked Diffusion Generation}

Unlike autoregressive models that decode from left to right, DLMs generate answers through iterative denoising \citep{sahoo2024simple,nie2025large,gao2026unifying}.
Let $Y^t=(Y_1^t,\ldots,Y_M^t)$ be the answer block at step $t$ and $U_t$ its unresolved masked positions.
Generation starts from a fully masked $Y^T$ with $U_T=\{1,\ldots,M\}$.
Given question $q$, visual evidence $E$, and state $Y^t$, one forward pass exposes $p_\theta^{(m)}(y\mid Y^t,q,E)$ for every $m\in U_t$.
A schedule selects $S_t\subseteq U_t$ and updates those positions from the same unresolved state using the position-wise proposal
\begin{equation}
Q_t(Y_{S_t}\mid Y^t,q,E)
=\prod_{m\in S_t}p_\theta^{(m)}(Y_m\mid Y^t,q,E).
\label{eq:dlm-preliminary}
\end{equation}
The process repeats until no masks remain.
Although implementations differ in their schedules and update budgets, their first forward pass exposes distributions over the prospective answer block before any answer token is committed.
This multi-position lookahead enables parallel decoding, but a position-wise proposal may fail to preserve a source choice shared across unresolved positions.
We analyze this first unresolved state and suppress the fixed $Y^T$ below.

\subsection{Problem Setting: Availability vs. Admission}

Each example consists of a question $q$, a visual corpus $\mathcal{D}$, and one or more reference answers.
A retriever returns a ranked candidate pool $C_k=(c_1,c_2,\ldots,c_k)$, from which the generator receives an admitted evidence sequence $E\subseteq C_k$.
Fixed top-$k$ input sets $E=C_k$, conflating candidate availability with evidence admission.
Answer-page Recall@$k$ measures availability---whether $C_k$ contains an annotated answer-bearing page---whereas answer accuracy measures generation quality under the admitted evidence $E$.
We seek to preserve the coverage benefit of a larger candidate pool while limiting harmful visual exposure during denoising.

\subsection{Source-Coherence Loss in Parallel Denoising}
\label{sec:source-theory}

The controlled interventions in Figure~\ref{fig:intro-semantic-conflict}(a) show that the degradation is driven primarily by competing visual semantics rather than the presence of a second visual slot.
We therefore model the retrieved pages as competing sources that support different coherent answers and formalize why this conflict is especially damaging when several positions are predicted from the same unresolved denoising state.

\begin{definition}[Conflict width and source coherence]
Let the correct and conflicting pages support answers $a=(a_1,\ldots,a_M)$ and $d=(d_1,\ldots,d_M)$.
The conflict set is $R=\{m:a_m\neq d_m\}$ and its width is $r=|R|$.
The answers $a$ and $d$ are \emph{source-coherent}; the hybrid set $\calH$ contains assignments that agree with them outside $R$ but select at least one token from each source inside $R$, and hence are supported by neither source.
\end{definition}

We formalize this unresolved source conflict through local two-source ambiguity and an exact-marginal factorized proposal on a restricted source-label process; their precise definitions and scope are given as Assumptions F.1 and F.2 in the Technical Supplement.
At each conflict position, this process retains and renormalizes the correct- and competing-source tokens.
Here $\epsilon\in(0,1)$ denotes competing-source mass in this restricted process.
The key issue is that positions in $R$ share a latent source choice, whereas the position-wise proposal does not.
For $r=2$, the mixture supports only $(a_1,a_2)$ and $(d_1,d_2)$, but the product assigns mass $2\epsilon(1-\epsilon)$ to the two cross-source combinations.

\begin{theorem}[Source-conflict amplification]
\label{thm:source-conflict}
Under Assumptions F.1 and F.2 in the Technical Supplement, for any $0<\epsilon<1$ and $r\geq1$, the factorized proposal assigns
\begin{equation}
\begin{aligned}
Q_{\parallel}(a\mid E)&=(1-\epsilon)^r, &
Q_{\parallel}(d\mid E)&=\epsilon^r,\\
Q_{\parallel}(\calH\mid E)&=1-(1-\epsilon)^r-\epsilon^r.
\end{aligned}
\label{eq:hybrid-main}
\end{equation}
Consequently, hybrid mass is zero for $r=1$ and strictly increases with conflict width $r$.
\end{theorem}

Thus, once several positions depend on the unresolved source, position-wise decisions can splice coherent branches and move probability mass to unsupported answers.

\begin{corollary}[Excess risk from lost source coupling]
\label{cor:parallel-risk}
Under Assumptions F.1 and F.2 in the Technical Supplement, suppose the coherent branch $a$ is the unique accepted assignment in this restricted process, and define its exact-match sampling risk as $\mathcal{R}(Q)=1-Q(a)$.
Let $Q_{\mathrm{causal}}$ be an exact causal factorization of the same mixture, which preserves the selected source after the prefix chooses a branch.
Then $\mathcal{R}_{\mathrm{causal}}=\epsilon$, and for every $r\geq2$,
\begin{equation}
\mathcal{R}_{\parallel}-\mathcal{R}_{\mathrm{causal}}
=(1-\epsilon)-(1-\epsilon)^r>0,
\qquad r\geq2.
\label{eq:risk-main}
\end{equation}
Hence factorizing exact marginals incurs excess risk solely from losing their shared source dependence.
\end{corollary}

The causal construction is an oracle comparator that isolates this dependence cost, rather than a universal ranking of model families.
The result explains why conflict is more damaging than a semantically empty input and motivates admission before final denoising; proofs are provided in the Technical Supplement.

\begin{figure}[t]
\centering
\includegraphics[width=\columnwidth]{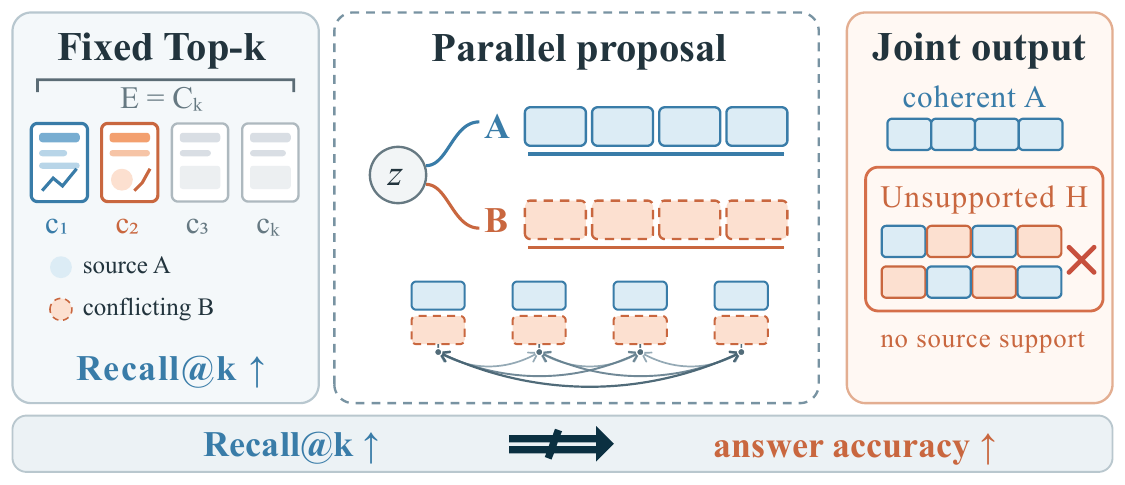}
\caption{Fixed top-$k$ input can improve answer-page availability while conflicting sources induce unsupported hybrid proposals.}
\label{fig:topk-failure}
\end{figure}

Figure~\ref{fig:topk-failure} summarizes the mechanism: fixed top-$k$ input can expose parallel denoising to competing visual sources, and the factorized position-wise proposal can convert unresolved source competition into unsupported hybrid answers.

\begin{table}[t]
\centering
\small
\setlength{\tabcolsep}{5pt}
\begin{tabular}{cccc}
\toprule
$r$ & LLaDA par. & LLaDA seq. & Qwen AR \\
\midrule
1 & 0.0000 & 0.0000 & 0.0000 \\
2 & 0.3017 & 0.1530 & $3.39\!\times\!10^{-7}$ \\
4 & 0.5842 & 0.2574 & $1.12\!\times\!10^{-8}$ \\
\bottomrule
\end{tabular}
\caption{Restricted hybrid source mass under controlled visual conflict. Each cell averages 32 base templates and both page orders; the parallel--sequential effect has the same direction under either order.}
\label{tab:hybrid-mass-main}
\end{table}

\subsection{Empirical Validation and Early Observability}

\paragraph{Controlled source-conflict validation.}
Table~\ref{tab:hybrid-mass-main} evaluates the source-conflict prediction through a same-weight LLaDA comparison differing only in proposal construction.
Hybrid mass is zero at $r=1$, increases with conflict width, and is consistently higher under parallel than source-conditioned sequential prediction for $r\geq2$, but remains negligible for native Qwen AR.
This supports source-coherence loss as the failure channel behind harmful multi-page input.

\paragraph{Early observability on natural retrieval.}
The same failure is visible before any answer token is committed.
On natural ChartQA cases, a conflicting page reduces first-step answer-block accuracy from 65.0 to 41.1 and produces the largest divergence from the correct-only distribution, whereas matched-blank and duplicated-correct controls remain substantially closer (Figure~\ref{fig:intro-semantic-conflict}(b)).
This ordering mirrors final-answer degradation (Figure~\ref{fig:intro-semantic-conflict}(a)), showing that conflict is already expressed in the initial answer state and persists through denoising.
The early-to-final correspondence motivates the first-step distribution as a generator-native signal for evidence admission.

\section{Entropy-Based Candidate Filter}

The analysis motivates \method{} to control visual evidence exposure along two axes: evidence granularity and final-context composition.
At evidence construction, full pages and layout-derived regions form a multi-granularity corpus spanning global and localized evidence.
At admission, \method{} evaluates candidates before decoding, using a geometry-matched blank to isolate their effect on prospective answer-block confidence.
Given a ranked top-$k$ pool, it expands the visual context only when supported by the DLM's pre-denoising state.

\begin{figure*}[t]
\centering
\includegraphics[width=\textwidth]{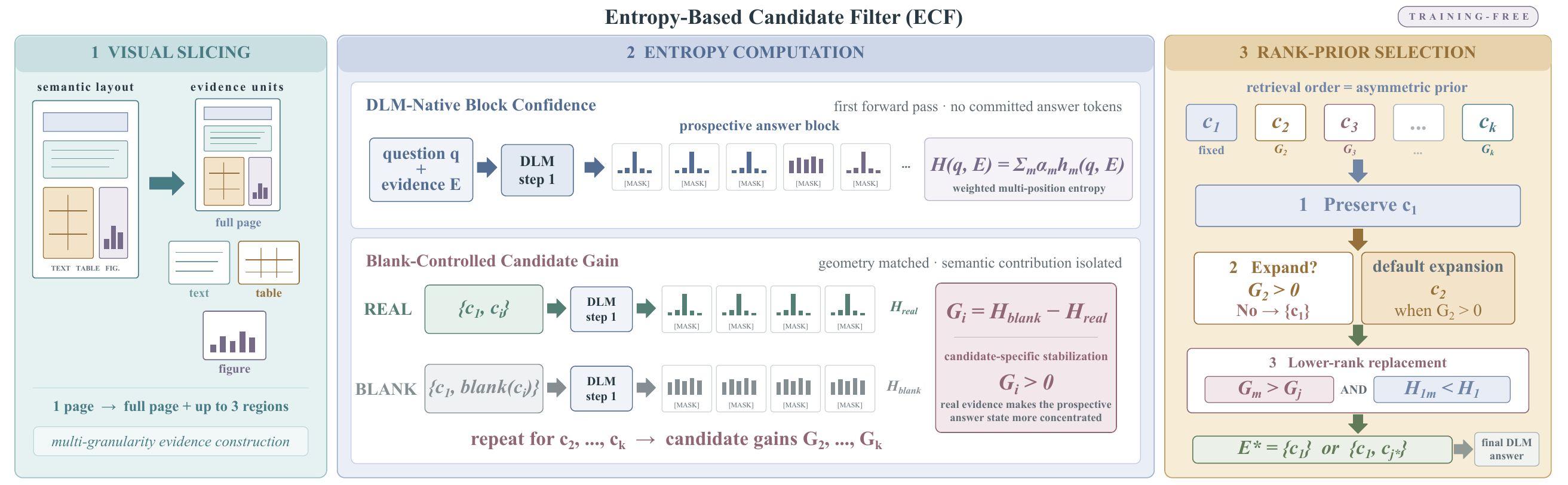}
\caption{Overview of ECF's three core components. Visual slicing constructs multi-granularity evidence units from full pages and layout-derived regions. Entropy computation measures prospective answer-block confidence and isolates each candidate's semantic contribution with a geometry-matched blank. Rank-prior selection combines these signals with retrieval order to preserve the top-ranked candidate and admit at most one companion.}
\label{fig:ecf-framework}
\end{figure*}

\subsection{DLM-Native Block Confidence}

Because candidate influence is visible before final denoising, we evaluate uncertainty in the DLM's prospective answer state.
For an evidence set $E$, define the entropy of answer slot $m$:
\begin{equation}
h_m(q,E)=-\sum_y p_\theta^{(m)}(y\mid q,E)
\log p_\theta^{(m)}(y\mid q,E).
\end{equation}
We aggregate the first answer block as
\begin{equation}
\begin{aligned}
H(q,E)&=\sum_{m=1}^{M}\alpha_m h_m(q,E),\\
\alpha_m&=\frac{\exp(-\lambda m)}{\sum_{j=1}^{M}\exp(-\lambda j)}.
\end{aligned}
\end{equation}
The decay rate $\lambda$ controls the emphasis on earlier answer positions; we use $\lambda=0.5$ throughout.
This block statistic uses the multi-position state implicated by Theorem~\ref{thm:source-conflict}, rather than relying on a single initial token that may encode formatting or a standard prefix.

\subsection{Blank-Controlled Candidate Gain}

Raw block-entropy changes mix the candidate's semantic contribution with the structural effect of adding another visual input.
\method{} separates these effects with geometry-matched real and blank-reference contexts.

For a top-ranked evidence unit $c_1$ and a candidate unit $c_i$, construct matched contexts
\begin{equation}
E_i^{\mathrm{real}}=\{c_1,c_i\}, \quad
E_i^{\mathrm{blank}}=\{c_1,\blank(c_i)\}.
\end{equation}
The blank has the same dimensions as $c_i$ and preserves the resulting image-token layout while removing candidate semantics.
We define the blank-controlled entropy gain
\begin{equation}
\gain_i = H(q,E_i^{\mathrm{blank}})-H(q,E_i^{\mathrm{real}}).
\end{equation}
The difference therefore isolates how $c_i$ changes the prospective answer state relative to an equally shaped semantic null.

\paragraph{Alignment with source-conflict risk.}

The matched-context decomposition and local correct-source condition are formalized in Assumption F.3 of the Technical Supplement.

\begin{proposition}[Local confidence--risk alignment]
\label{prop:entropy-risk}
Under Assumption F.3 in the Technical Supplement, for $r\geq2$, positive candidate gain is equivalent to reducing competing-source mass, hybrid mass, and parallel exact-match risk:
\begin{equation}
\begin{aligned}
\gain_i>0
&\Longleftrightarrow
\epsilon_i<\epsilon_{\blank}\\
&\Longleftrightarrow
\Phi_r(\epsilon_i)<\Phi_r(\epsilon_{\blank})\\
&\Longleftrightarrow
\mathcal{R}_{\parallel}(\epsilon_i)
<\mathcal{R}_{\parallel}(\epsilon_{\blank}),
\end{aligned}
\label{eq:entropy-risk-alignment}
\end{equation}
where $\epsilon_i$ and $\epsilon_{\blank}$ are the competing-source masses under the real and blank contexts, respectively, $\Phi_r(\epsilon)=1-(1-\epsilon)^r-\epsilon^r$ is the hybrid mass, and
$\mathcal{R}_{\parallel}(\epsilon)=1-(1-\epsilon)^r$ is the parallel exact-match risk.
\end{proposition}

Proposition~\ref{prop:entropy-risk} gives the gain a direct operational meaning: relative to an equally shaped blank, positive gain indicates that the candidate reduces competing-source uncertainty and therefore lowers both hybrid mass and parallel sampling risk.
This is why \method{} uses a candidate-specific entropy difference rather than raw entropy.
The alignment is deliberately local: outside the correct-source basin, low entropy may reflect a confidently wrong source, so retrieval rank supplies the complementary prior.

\begin{table*}[t]
\centering
\small
\setlength{\tabcolsep}{7pt}
\begin{tabular}{lccccc}
\toprule
Method & ChartQA & InfoChartQA & DocVQA & InfoVQA & TATDQA \\
\midrule
Fixed top-2 & 24.60 & 18.96 & 30.88 & 30.12 & 11.52 \\
Frozen Semantic Top-2 & 25.88 & 17.16 & 31.64 & 32.80 & 12.16 \\
SPREAD Top-2 & 26.12 & 18.68 & 33.56 & 32.72 & 12.60 \\
SARDI Top-2 & 27.20 & 19.04 & 36.12 & 34.88 & 12.88 \\
UOVES Top-2 & 26.84 & 20.00 & \textbf{37.16} & 35.92 & 13.76 \\
Answer-UQ Top-1 & \underline{30.72} & \underline{20.68} & 32.92 & \underline{36.08} & \underline{14.08} \\
\method{} & \textbf{31.12} & \textbf{21.16} & \underline{36.72} & \textbf{42.60} & \textbf{18.96} \\
\bottomrule
\end{tabular}
\caption{Answer accuracy (\%) under fixed top-$k$ input and training-free evidence-selection alternatives with LLaDA2.0-Uni. Fixed top-2 input unconditionally passes the two highest-ranked retrieved pages; \method{} and the other baselines use a candidate-pool size of $k=3$ and admit at most one companion.}
\label{tab:training-free}
\end{table*}

\begin{table*}[t]
\centering
\small
\setlength{\tabcolsep}{3.5pt}
\begin{tabular}{llccccc@{\hspace{11pt}}c}
\toprule
Model & Input & ChartQA & InfoChartQA & DocVQA & InfoVQA & TATDQA & Avg. \\
\midrule
\multirow{4}{*}{LLaDA2.0-Uni}
& fixed top-1 & \underline{29.96} & \underline{20.44} & 27.32 & \underline{30.20} & 11.60 & 23.90 \\
& fixed top-2 & 24.60 & 18.96 & \underline{30.88} & 30.12 & 11.52 & 23.22 \\
& fixed top-3 & 21.20 & 16.40 & 30.12 & 27.44 & \underline{12.20} & 21.47 \\
& \method{} & \textbf{31.12} & \textbf{21.16} & \textbf{36.72} & \textbf{42.60} & \textbf{18.96} & \textbf{30.11} \\
\midrule
\multirow{4}{*}{LLaDA-V}
& fixed top-1 & \textbf{22.28} & \textbf{17.12} & \underline{22.68} & \underline{26.48} & \underline{9.88} & 19.69 \\
& fixed top-2 & 13.32 & 12.60 & 15.88 & 18.32 & 8.72 & 13.77 \\
& fixed top-3 & 10.04 & 9.00 & 11.36 & 12.92 & 7.40 & 10.14 \\
& \method{} & \underline{20.48} & \underline{16.28} & \textbf{24.44} & \textbf{30.84} & \textbf{14.04} & \textbf{21.22} \\
\midrule
\multirow{4}{*}{Dream-VL}
& fixed top-1 & 31.92 & 23.12 & 34.52 & 44.72 & \underline{17.08} & 30.27 \\
& fixed top-2 & 33.60 & \underline{25.32} & \textbf{38.48} & \underline{47.64} & 17.00 & 32.41 \\
& fixed top-3 & \textbf{34.44} & \textbf{25.48} & 35.40 & 44.92 & 17.00 & 31.45 \\
& \method{} & \underline{34.40} & 25.08 & \underline{37.60} & \textbf{49.68} & \textbf{21.16} & \textbf{33.58} \\
\bottomrule
\end{tabular}
\caption{Answer accuracy (\%) across DLM backends. \method{} uses candidate-pool size $k=3$. For each dataset within a backend, the best and second-best results are shown in bold and underlined, respectively; only the best average is shown in bold.}
\label{tab:cross-dlm}
\end{table*}

\begin{table*}[t]
\centering
\small
\setlength{\tabcolsep}{4pt}
\begin{tabular}{llccccc}
\toprule
Source & Metric & ChartQA & InfoChartQA & DocVQA & InfoVQA & TATDQA \\
\midrule
\multirow{3}{*}{VisRAG pool}
& Recall@1 & 34.08 & 38.16 & 34.04 & 51.08 & 15.76 \\
& Recall@2 & 41.68 & 46.40 & 43.64 & 62.04 & 22.68 \\
& Recall@3 & 46.00 & 50.08 & 48.36 & 68.08 & 27.88 \\
\midrule
LLaDA2.0-Uni + \method{} & Answer-page usage & 39.60 & 42.28 & 39.20 & 59.36 & 24.32 \\
LLaDA-V + \method{} & Answer-page usage & 39.32 & 42.32 & 40.92 & 59.80 & 25.16 \\
Dream-VL + \method{} & Answer-page usage & 38.80 & 41.44 & 40.56 & 60.32 & 25.24 \\
\bottomrule
\end{tabular}
\caption{Answer-page Recall@$k$ and answer-page usage (\%).}
\label{tab:retrieval-usage}
\end{table*}

\begin{table}[t]
\centering
\small
\begin{tabular*}{\columnwidth}{@{\extracolsep{\fill}}lcccc@{}}
\toprule
Dataset & $k$ & Acc. (\%) & Recall@$k$ (\%) & Page use (\%) \\
\midrule
\multirow{2}{*}{ChartQA} & 4 & 31.52 & 48.64 & 39.84 \\
 & 5 & 31.80 & 51.20 & 39.92 \\
\midrule
\multirow{2}{*}{InfoChartQA} & 4 & 21.20 & 52.64 & 42.40 \\
 & 5 & 20.96 & 54.80 & 42.08 \\
\midrule
\multirow{2}{*}{DocVQA} & 4 & 36.72 & 49.04 & 39.24 \\
 & 5 & 37.04 & 51.96 & 39.52 \\
\midrule
\multirow{2}{*}{InfoVQA} & 4 & 42.78 & 70.43 & 59.50 \\
 & 5 & 43.02 & 73.35 & 59.06 \\
\midrule
\multirow{2}{*}{TATDQA} & 4 & 18.72 & 37.00 & 23.28 \\
 & 5 & 18.76 & 40.12 & 24.60 \\
\bottomrule
\end{tabular*}
\caption{\method{} with larger candidate pools.}
\label{tab:larger-k}
\end{table}

\subsection{Rank-Prior Selection}

The gain provides generator-side evidence for each candidate, while retrieval order supplies the prior for searching the ranked pool.
\method{} combines both signals so that lower-ranked candidates replace the default expansion only with stronger evidence.
It consumes only the ranked candidates, without retriever-specific scores, score calibration, or learned parameters, and can be paired with any visual retriever that returns an ordered list.

\begin{algorithm}[!t]
\caption{Rank-Prior Selection in \method{}}
\label{alg:ecf}
\begin{algorithmic}[1]
\REQUIRE Question $q$; ranked candidate pool $C_k=(c_1,\ldots,c_k)$; DLM $p_\theta$; matched-blank operator $\blank(\cdot)$
\ENSURE Admitted evidence set $E^\star$
\STATE \textit{Stage 1: Preserve the top-ranked candidate}
\IF{$k=1$}
    \STATE \textbf{return} $\{c_1\}$
\ENDIF
\STATE \textit{Stage 2: Test the default rank-2 expansion}
\STATE $H_1 \gets H(q,\{c_1\})$
\STATE $\gain_2 \gets H(q,E_2^{\mathrm{blank}})-H(q,E_2^{\mathrm{real}})$
\IF{$\gain_2 \leq 0$}
    \STATE \textbf{return} $\{c_1\}$
\ENDIF
\STATE $j \gets 2$
\STATE \textit{Stage 3: Evaluate lower-ranked replacements}
\FOR{$m=3,\ldots,k$}
    \STATE $\gain_m \gets H(q,E_m^{\mathrm{blank}})-H(q,E_m^{\mathrm{real}})$
    \STATE $H_{1m} \gets H(q,\{c_1,c_m\})$
    \IF{$\gain_m>\gain_j$ and $H_{1m}<H_1$}
        \STATE $j \gets m$
    \ENDIF
\ENDFOR
\STATE \textit{Stage 4: Construct the admitted evidence set}
\STATE \textbf{return} $\{c_1,c_j\}$
\end{algorithmic}
\end{algorithm}

The policy always retains $c_1$ and admits at most one companion.
For $k\geq2$, $c_2$ serves as the rank-prior default and is admitted only if $\gain_2>0$; otherwise, the policy returns $E^\star=\{c_1\}$.
Conditional on $\gain_2>0$, the admissible lower-ranked replacements are
\begin{equation}
\begin{aligned}
\mathcal{A}_k=\{m\in\{3,\ldots,k\}:{}&\gain_m>\gain_2,\\[-2pt]
&H(q,\{c_1,c_m\})<H(q,\{c_1\})\}.
\end{aligned}
\end{equation}
The selected companion is
\begin{equation}
j^\star=\arg\max_{j\in\{2\}\cup\mathcal{A}_k}\gain_j,
\qquad E^\star=\{c_1,c_{j^\star}\},
\label{eq:topk-policy}
\end{equation}
with ties favoring the higher-ranked candidate.
This construction treats $c_2$ as the default expansion while requiring lower-ranked candidates to improve both relative gain and absolute joint confidence.
Its probe cost grows linearly with $k$, but each real or blank probe requires only a single first-step forward pass rather than a complete denoising trajectory.
Full answer generation is performed only once with at most two admitted images.

\section{Experiments}

\subsection{Experimental Setup}

\paragraph{Benchmarks and system setup.}
We evaluate \method{} with three multimodal DLM backends: LLaDA2.0-Uni \citep{inclusionai2026llada2uni}, LLaDA-V \citep{you2025lladav}, and Dream-VL \citep{ye2025dreamvl}.
For retrieval, a VisRAG-style retriever \citep{yu2024visrag} ranks visual evidence units for each question.

We evaluate five visual QA benchmarks: ChartQA \citep{masry2022chartqa} and InfoChartQA \citep{xie2025infochartqa} for charts, plus DocVQA \citep{mathew2021docvqa}, InfoVQA \citep{mathew2022infographicvqa}, and TATDQA \citep{zhu2022tatdqa} for dense documents. \method{} retrieves full pages and layout-derived slices from a multi-granularity corpus.

\paragraph{Evaluation protocol.}
We report answer accuracy (\%) as the primary metric, using a unified relaxed matching rule for textual and numeric answers across all datasets and models.
For diagnostics, we also report Recall@$k$ for answer-page availability and answer-page usage for the presence of answer-bearing evidence in the admitted context.
Further details appear in the Technical Supplement.

\paragraph{Training-Free Alternatives.}
We compare \method{} with five training-free alternatives under LLaDA2.0-Uni.
UOVES selects two pages with a multimodal surrogate; Answer-UQ selects the candidate with the lowest generated-answer uncertainty; Frozen Semantic reranks with a frozen vision--text encoder; SPREAD changes reveal order using query relevance; and SARDI retrieves from tentative DLM predictions \citep{luo2026uoves,yu2026spread,junger2026sardi}.
All methods share VisRAG retrieval, the answer budget, and the scorer; fixed top-$k$ inputs expand evidence unconditionally.

\subsection{Main Results}

\paragraph{LLaDA2.0-Uni.}

Table~\ref{tab:training-free} tests whether \method{}'s gains merely reflect avoiding unconditional evidence expansion.
\method{} improves over matched fixed top-2 input by 6.90 percentage points on average.
More importantly, it retains a 2.37-percentage-point average margin over the strongest training-free alternative on each dataset, even though these methods already select, rerank, or adapt retrieved evidence.
Thus, passing fewer pages alone does not explain the gain, supporting a generator-native criterion for evidence admission before denoising.

\paragraph{Cross-DLM validation.}

Table~\ref{tab:cross-dlm} evaluates the admission rule across DLM backends.
\method{} achieves the best accuracy in 10 of the 15 model--dataset pairs and the highest average for all three backends.
Across the 15 pairs, the same training-free rule improves accuracy by 2.62 percentage points on average over the strongest fixed top-$k$ input, without backend-specific training or calibration.
Its cross-backend behavior characterizes \method{} as a reusable evidence-admission layer rather than a generator-specific optimization.

\paragraph{Candidate Availability and Answer-Page Usage.}

Table~\ref{tab:retrieval-usage} separates candidate availability from evidence admitted for generation.
Recall@$k$ grows with pool size, but Table~\ref{tab:cross-dlm} shows that admitting more pages does not correspondingly improve accuracy.
Answer-page usage measures whether answer-bearing evidence survives admission, separating retrieval coverage from effective generation context.

\subsection{Analysis and Ablations}
\label{sec:ablations}

\paragraph{Larger candidate pools.}

Table~\ref{tab:larger-k} evaluates \method{} beyond the main $k=3$ candidate pool.
Unlike fixed top-$k$ input, whose accuracy decreases as more retrieved pages enter the decoding context (Figure~\ref{fig:intro-retrieval-accuracy}), \method{} remains stable as its candidate pool grows: macro accuracy is 30.11, 30.19, and 30.32 for $k=3,4,5$, respectively, despite diminishing marginal Recall gains.
Increasing $k$ expands only the candidates screened by \method{}, while final decoding remains capped at two images.
This decoupling makes $k$ a flexible retrieval budget: small pools suffice when retrieval saturates early, whereas larger pools can exploit informative lower-ranked evidence from weaker retrievers or more complex datasets without the degradation of fixed top-$k$ input.
\method{} thus adapts across retrieval regimes rather than relying on a fixed $k$.

\begin{figure}[t]
\centering
\includegraphics[width=\columnwidth]{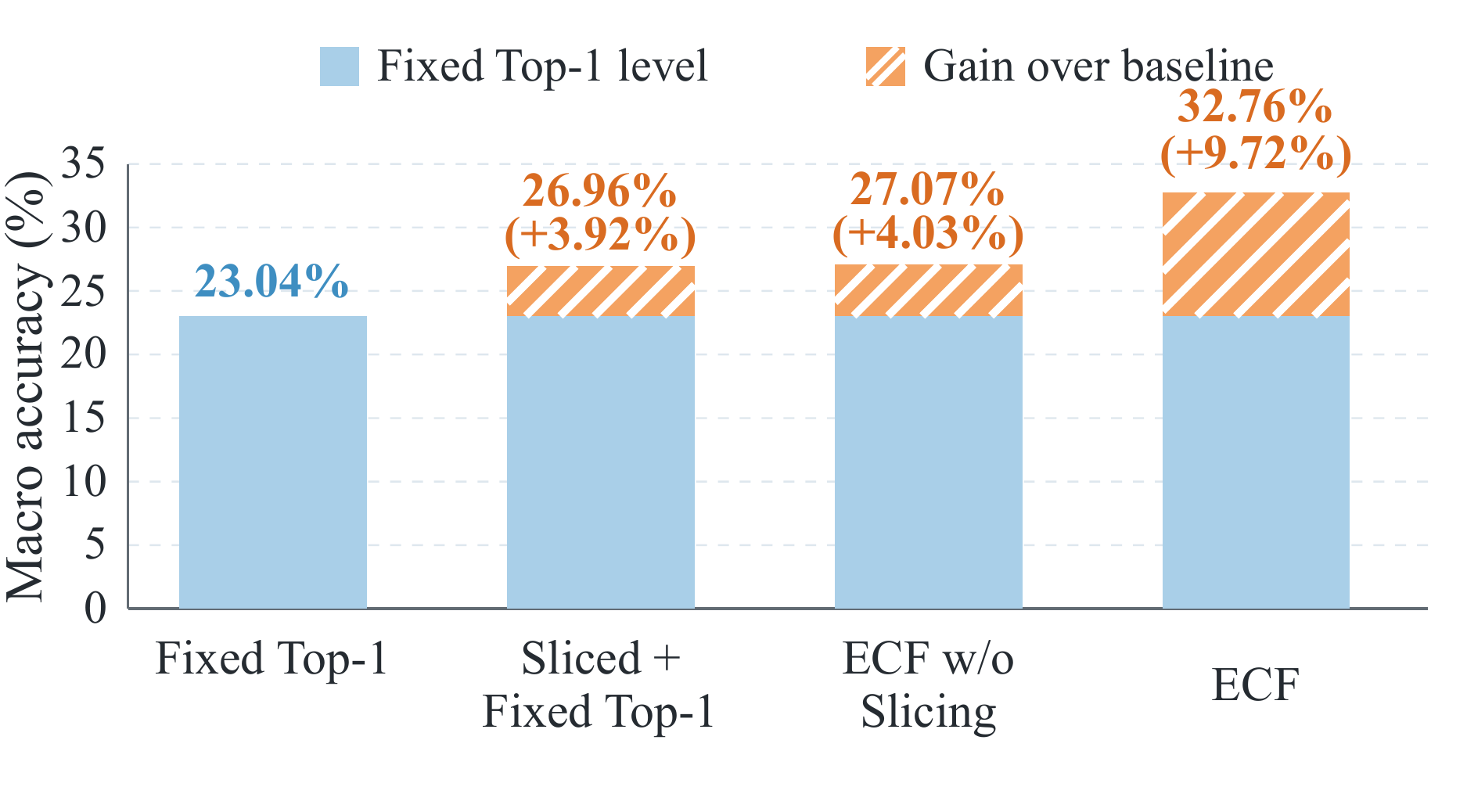}
\caption{Macro-average answer accuracy across DocVQA, InfoVQA, and TATDQA for the slicing and gating ablation.}
\label{fig:slicing-gating}
\end{figure}

\paragraph{Slicing and admission components.}

Figure~\ref{fig:slicing-gating} isolates the contributions of the two components in the complete \method{} pipeline: layout-aware slicing and entropy-based candidate filtering.
Each component independently improves the fixed top-1 baseline by about four percentage points, while combining them yields a substantially larger gain of 9.72 percentage points.
These gains reflect their complementary roles: slicing reduces unrelated content within each retrieved image, whereas admission limits interference across images in the final denoising context.

\section{Conclusion}

Retrieving more evidence does not guarantee better visual DLM-RAG: larger candidate pools improve answer-page availability, but admitting every candidate can reduce accuracy when position-wise parallel proposals lose shared visual-source dependence and assign probability to unsupported answers.
Controlled and natural-retrieval studies locate this interference in the first-step answer block, enabling evidence assessment before decoding.
\method{} combines multi-granularity evidence construction with blank-controlled, rank-aware admission to reduce within-candidate noise and limit which candidates reach final denoising.
Across five visual QA benchmarks and three multimodal DLMs, \method{} averages a 2.62-percentage-point gain over the strongest fixed top-$k$ input and remains stable as candidate pools grow.
These findings recast pool size as a retrieval budget: broader retrieval helps when evidence is screened before denoising rather than accumulated unconditionally.

\bibliography{references}

%% file: technical_supplement.tex
\maketitle

\appendix

\section{Appendix A: Implementation Details}

\paragraph{Evaluation subset.}
To ensure a consistent evaluation scale across datasets, we randomly sample 2,500 question--answer pairs from the test split of each of the five datasets whenever available.
For each model--dataset--configuration combination, we report one complete evaluation run over the corresponding fixed evaluation subset.

\paragraph{Prompt wrapper.}
All models use the same task-level wrapper:
\begin{quote}
\texttt{Question: <question and}\\
\texttt{short-answer instruction>}\\
\texttt{Answer:}
\end{quote}
Model-specific image tokens and chat templates are treated as unavoidable backend differences.

\paragraph{Answer length.}
All backends generate an eight-token short-answer region.
LLaDA2.0-Uni uses its native block-denoising implementation with 32 denoising updates for the answer block.
LLaDA-V and Dream-VL use their official full-answer denoising interfaces with eight update steps.
The same answer length is used for entropy probing and final generation within each backend.

\section{Appendix B: Unified Answer Scoring}
\label{app:scoring}

The same scorer is used across all datasets and model backends.
Predictions and references are normalized by extracting the answer content, removing prefixes, suffixes, symbols, and whitespace, and converting letters to lowercase.
Non-numeric answers are evaluated by exact or constrained substring matching, while numeric answers use exact or bounded edit-distance matching.
When multiple references are available, matching any reference is considered correct.

\section{Appendix C: Layout-Aware Visual Slicing}
\label{app:slicing}

We apply pre-retrieval slicing to DocVQA, InfoVQA, and TATDQA.
ChartQA and InfoChartQA use the original image corpus without slicing.
The splitter uses the \texttt{surya-ocr} layout predictor to detect document regions \citep{paruchuri2025surya}.
Detected boxes receive eight pixels of padding and regions smaller than 3\% of the page area or with a side shorter than 96 pixels are iteratively merged with their nearest region.
If more than three regions remain, the smallest regions are merged until at most three fragments remain.
The fragments are ordered by their page coordinates.

For each original page, the retrieval corpus contains all retained fragments together with the uncropped complete page.
Consequently, one source page contributes at most four candidate images: up to three fragments and one full-page image.
This construction preserves global context while allowing VisRAG and \method{} to operate on finer-grained visual evidence.

Relative to the standard full-image inputs reported in the main paper, the multi-granularity corpus improves every matched fixed-$k$ setting by 1.40--9.24 percentage points.
Nevertheless, the complete \method{} pipeline remains 1.72--4.00 percentage points higher than the strongest sliced-corpus fixed top-$k$ result on each dataset, showing that localized candidate construction and selective admission are complementary.

\section{Appendix D: Answer-Page Recall and Usage Metrics}
\label{app:retrieval-success}

Recall@$k$ is successful when at least one of the top-$k$ retrieved candidates maps to an answer-bearing source page.
For an unsliced candidate, the mapping is the candidate page itself.
For a sliced candidate, every fragment stores its parent-page identifier; retrieving any fragment whose parent is an answer-bearing page is counted as successful retrieval of that page.
The same parent mapping is used for \method{}'s final answer-page usage metric.
Thus a final evidence sequence is successful if it contains either the complete answer page or any fragment derived from that page.

\section{Appendix E: Detailed Candidate-Admission Analyses}
\label{app:larger-k}

\subsection*{Slicing and Admission Components}

Table~\ref{tab:slicing-gating-detailed} reports the per-dataset results underlying the macro-average slicing and admission ablation in the main paper.

The full-page comparison isolates confidence-based admission, whereas the sliced-plus-full-page setting evaluates the complete pipeline.
Together with Table~\ref{tab:sliced-topk}, these results separate candidate granularity from evidence admission.

\begin{table}[!ht]
\centering
\small
\setlength{\tabcolsep}{2.5pt}
\begin{tabular}{llccc}
\toprule
Corpus & Selector & DocVQA & InfoVQA & TATDQA \\
\midrule
Full page & Fixed top-1 & 27.32 & 30.20 & 11.60 \\
Full page & \method{} & 35.52 & 35.08 & 10.60 \\
Sliced + full page & Fixed top-1 & 28.72 & 36.04 & 16.12 \\
Sliced + full page & \method{} & \textbf{36.72} & \textbf{42.60} & \textbf{18.96} \\
\bottomrule
\end{tabular}
\caption{Per-dataset answer accuracy (\%) underlying the macro-average slicing and admission ablation in the main paper.}
\label{tab:slicing-gating-detailed}
\end{table}

\begin{table}[!ht]
\centering
\small
\setlength{\tabcolsep}{5pt}
\begin{tabular}{lccc}
\toprule
Dataset & Top-1 & Top-2 & Top-3 \\
\midrule
DocVQA & 28.72 & 32.76 & 33.76 \\
InfoVQA & 36.04 & 38.60 & 36.68 \\
TATDQA & 16.12 & 17.16 & 17.24 \\
\bottomrule
\end{tabular}
\caption{Answer accuracy (\%) of LLaDA2.0-Uni with fixed top-$k$ input from the same sliced-plus-full-page candidate corpus used by \method{}.}
\label{tab:sliced-topk}
\end{table}

\providecommand{\beginlargerpooltable}{\begin{table*}[!t]}
\newcommand{\renderlargerpooltable}{%
\beginlargerpooltable
\centering
\small
\setlength{\tabcolsep}{4pt}
\begin{tabular}{lccccccccc}
\toprule
Dataset & $k$ & Acc. (\%) & Recall@$k$ (\%) & Page use (\%) & Admit (\%)
& Rank 2 (\%) & Rank 3 (\%) & Rank 4 (\%) & Rank 5 (\%) \\
\midrule
\multirow{2}{*}{ChartQA} & 4 & 31.52 & 48.64 & 39.84 & 48.96 & 31.60 & 9.08 & 8.28 & -- \\
 & 5 & 31.80 & 51.20 & 39.92 & 48.96 & 27.44 & 7.64 & 6.60 & 7.28 \\
\midrule
\multirow{2}{*}{InfoChartQA} & 4 & 21.20 & 52.64 & 42.40 & 50.08 & 33.48 & 8.92 & 7.68 & -- \\
 & 5 & 20.96 & 54.80 & 42.08 & 50.08 & 28.12 & 7.88 & 6.44 & 7.64 \\
\midrule
\multirow{2}{*}{DocVQA} & 4 & 36.72 & 49.04 & 39.24 & 47.76 & 30.28 & 8.36 & 9.12 & -- \\
 & 5 & 37.04 & 51.96 & 39.52 & 47.76 & 25.64 & 7.08 & 7.92 & 7.12 \\
\midrule
\multirow{2}{*}{InfoVQA} & 4 & 42.78 & 70.43 & 59.50 & 52.22 & 33.97 & 9.16 & 9.08 & -- \\
 & 5 & 43.02 & 73.35 & 59.06 & 52.18 & 29.01 & 8.04 & 7.96 & 7.16 \\
\midrule
\multirow{2}{*}{TATDQA} & 4 & 18.72 & 37.00 & 23.28 & 47.28 & 33.28 & 7.16 & 6.84 & -- \\
 & 5 & 18.76 & 40.12 & 24.60 & 47.28 & 29.16 & 6.08 & 5.72 & 6.32 \\
\bottomrule
\end{tabular}
\ifdefined\combinedappendixlayout
  \addtocounter{table}{3}
\fi
\caption{LLaDA2.0-Uni \method{} results with larger candidate pools. Accuracy, Recall@$k$, answer-page usage, total admission rate, and selected-rank rates are percentages. Rank-use columns sum to the total admission rate up to rounding.}
\label{supp:tab:larger-k}
\ifdefined\combinedappendixlayout
  \addtocounter{table}{-4}
\fi
\end{table*}}
\ifdefined\combinedappendixlayout
  \renderlargerpooltable
\fi

\subsection*{Rank-Prior Selection}

Rank-prior selection combines retrieval order as an asymmetric prior with counterfactual entropy gain as generator-side evidence.
A positive $\gain_2$ shows that candidate-specific information stabilizes the top-1-conditioned answer state relative to matched placeholder expansion, providing evidence that expansion is warranted.
Because marginal answer-page recall decreases with retrieval rank, $c_2$ remains the default admitted candidate.
Any lower-ranked candidate must exceed the current best gain and reduce joint entropy relative to top-1 before replacing that default.
Table~\ref{tab:rank-policy} compares the main rank-prior selection policy at $k=3$ with two-candidate \method{} at $k=2$ and direct maximum-gain selection at $k=3$.

\begin{table}[!ht]
\centering
\small
\setlength{\tabcolsep}{3pt}
\begin{tabular}{lccc}
\toprule
Policy & Acc. (\%) & Page use (\%) & Rank 3 (\%) \\
\midrule
\method{} ($k=2$) & 30.60 & 38.92 & -- \\
Direct max-gain ($k=3$) & 30.08 & 40.28 & 32.16 \\
\method{} ($k=3$) & \textbf{31.12} & 39.60 & 11.04 \\
\bottomrule
\end{tabular}
\caption{Candidate-rank policy on ChartQA with LLaDA2.0-Uni. Direct max-gain omits the rank-prior replacement rule, while \method{} ($k=2$) isolates two-candidate admission. Page use is answer-page usage, and Rank 3 is the rank-3 selection rate.}
\label{tab:rank-policy}
\end{table}

The asymmetric policy retains the larger marginal retrieval value of rank 2 while allowing rank 3 to replace it when both the relative counterfactual gain and absolute top-1 comparison are favorable.
Compared with direct max-gain selection, it changes the selected rank on 528 examples and improves answer accuracy by 1.04 percentage points.

\subsection*{Blockwise Entropy Aggregation}

Table~\ref{tab:entropy-aggregation} tests whether the entire masked answer block provides a more informative confidence statistic than a first-position view.

\begin{table}[!ht]
\centering
\small
\setlength{\tabcolsep}{3pt}
\begin{tabular}{lccc}
\toprule
Confidence statistic & Page use (\%) & Acc. (\%) & Admit (\%) \\
\midrule
First mask entropy & 38.72 & 29.76 & 50.24 \\
Uniform block mean & \textbf{38.96} & 30.52 & 49.28 \\
Weighted block mean & 38.92 & \textbf{30.60} & 48.96 \\
\bottomrule
\end{tabular}
\caption{First-step entropy aggregation with $k=2$ on ChartQA with LLaDA2.0-Uni. Page use denotes answer-page usage, and Admit is the percentage of examples using two images.}
\label{tab:entropy-aggregation}
\end{table}

To isolate the confidence statistic from lower-rank replacement, all variants use $k=2$, so the only decision is whether $c_2$ should accompany $c_1$.
The weighted block statistic increases answer-page usage by 0.20 percentage points and answer accuracy by 0.84 percentage points over first-mask entropy.
Although modest, the consistent improvement supports using the DLM's multi-position lookahead rather than tying admission to a single slot that may encode formatting or a standardized prefix.

\subsection*{Counterfactual Visual Controls}

Directly comparing $H(q,c_1)$ and $H(q,c_1,c_i)$ is not a clean evidence test because the two contexts differ in visual input count and model-specific image-token layout.
The main method therefore uses a white image with the candidate's dimensions to preserve input geometry while removing visual semantics.
This makes $\gain_i$ a more specific estimate of whether the candidate evidence unit itself stabilizes denoising.
To isolate reference construction from lower-rank replacement, all control variants use $k=2$.
Table~\ref{tab:control-ablation} compares this choice with two alternatives that retain semantic or low-level visual content.

\begin{table}[!ht]
\centering
\small
\setlength{\tabcolsep}{3pt}
\begin{tabular}{lcc}
\toprule
Control & Acc. (\%) & Admit (\%) \\
\midrule
Patch shuffle & 29.56 & 55.60 \\
Unrelated real page & 29.12 & 53.40 \\
White blank & \textbf{30.60} & 48.96 \\
\bottomrule
\end{tabular}
\caption{Counterfactual visual controls with $k=2$ on ChartQA. Accuracy and admission rate are percentages.}
\label{tab:control-ablation}
\end{table}

The white blank gives the highest answer accuracy and the lowest admission rate among the three controls.
Unlike patch shuffling, which preserves page-specific appearance, and an unrelated real page, which introduces new semantics, it removes candidate semantics while preserving input geometry; \method{} therefore uses it as the matched control.

\subsection*{Larger Candidate Pools and Rank Use}

Table~\ref{supp:tab:larger-k} extends the LLaDA2.0-Uni \method{} setting beyond the main $k=3$ candidate pool.

\ifdefined\combinedappendixlayout
  \stepcounter{table}
\else
  \renderlargerpooltable
\fi

Increasing $k$ raises retrieval coverage, and \method{} remains able to exploit the additional candidates when they provide useful confidence gains.
The gains are not purely monotonic, because larger pools are only helpful when the retriever places answer-bearing evidence in the added ranks and the admission mechanism can separate it from conflicting context.
This supports treating $k$ as a flexible candidate-availability parameter: stronger or broader retrieval can make larger pools useful, while final generation remains bounded by the admission rule.
\FloatBarrier

\section{Appendix F: Formal Assumptions and Proofs for the Source-Conflict Analysis}
\label{app:theory-proof}

\subsection*{Scope and Restricted Source Process}

We analyze the first prospective answer-block distribution before any answer
token is committed.  The analysis concerns a restricted source process, not
the DLM's complete full-vocabulary output law or its entire iterative
denoising trajectory.  Fix a question $q$, admitted visual evidence $E$, and
two source-coherent token sequences $a=(a_1,\ldots,a_M)$ and
$d=(d_1,\ldots,d_M)$ supported by the correct and competing visual sources.
Their conflict set and conflict width are
\begin{equation}
R=\{m:a_m\neq d_m\}, \qquad r=|R|.
\end{equation}
At each $m\in R$, define a source label $Z_m\in\{A,D\}$, where $A$ maps to
$a_m$ and $D$ maps to $d_m$.  Operationally, the controlled study obtains the
corresponding position-wise source probabilities by restricting the model
distribution to these two tokens and renormalizing.
The source-coherent label vectors are $A^r$ and $D^r$; the hybrid set
$\calH_R=\{A,D\}^r\setminus\{A^r,D^r\}$ contains vectors that select at
least one token from each source.  This is the same restricted source mass
measured in the controlled validation.  Probability assigned by the full
model to tokens outside each source-token pair is intentionally outside this
local abstraction.

\paragraph{Assumption F.1 (Unresolved two-source ambiguity).}
Before the visual source is resolved, a shared latent variable
$S\in\{A,D\}$ selects the correct or competing source with probabilities
$1-\epsilon$ and $\epsilon$, respectively.  Conditioned on $S$, all labels
in $R$ follow the same source, so the restricted joint law is
\begin{equation}
\begin{aligned}
P_R(Z_R\mid E)
&=(1-\epsilon)\,\delta_{A^r}(Z_R)
+\epsilon\,\delta_{D^r}(Z_R),\\
&\hspace{35mm}0<\epsilon<1.
\end{aligned}
\label{eq:supp-source-mixture}
\end{equation}
Here $\epsilon$ is the competing-source mass.  Assumption F.1 specifies only
the normalized source-label process on $R$; it does not assert that the
full-vocabulary model distribution has support only on $a$ and $d$.

\paragraph{Assumption F.2 (Exact-marginal factorized proposal).}
To isolate dependence loss rather than marginal prediction error, the first
prospective simultaneous proposal uses the exact marginals of
Eq.~\eqref{eq:supp-source-mixture} but factorizes them:
\begin{equation}
Q_{\parallel,R}(Z_R\mid E)
=\prod_{m\in R}P_R(Z_m\mid E).
\label{eq:supp-factorized-proposal}
\end{equation}
Thus $Q_{\parallel,R}(Z_m=A\mid E)=1-\epsilon$ and
$Q_{\parallel,R}(Z_m=D\mid E)=\epsilon$ at every $m\in R$.
This marginally exact construction isolates dependence loss; Lemma F.1 below
further shows that it is the forward-KL-optimal product approximation.  It
models the restricted source-label projection of the prospective block exposed
by the first forward pass; it is not identified with the full-vocabulary block
or the final distribution produced by an arbitrary multi-step denoising
schedule.

To match the compact notation in the main paper, within this local model we
write
\begin{equation}
\begin{aligned}
Q_{\parallel}(a\mid E)&:=Q_{\parallel,R}(A^r\mid E),\\
Q_{\parallel}(d\mid E)&:=Q_{\parallel,R}(D^r\mid E),\\
Q_{\parallel}(\calH\mid E)&:=Q_{\parallel,R}(\calH_R\mid E).
\end{aligned}
\label{eq:supp-notation-bridge}
\end{equation}
The probabilities in the main paper's theorem and corollary therefore refer
to this restricted source process, not to unconditional full-vocabulary
sequence probability.

\subsection*{Theorem 1: Source-Conflict Amplification}

\noindent\textbf{Theorem 1.}
Under Assumptions F.1 and F.2, for any $0<\epsilon<1$ and $r\geq1$,
\begin{equation}
\begin{aligned}
Q_{\parallel,R}(A^r\mid E)&=(1-\epsilon)^r,\\
Q_{\parallel,R}(D^r\mid E)&=\epsilon^r,\\
Q_{\parallel,R}(\calH_R\mid E)
&=1-(1-\epsilon)^r-\epsilon^r.
\end{aligned}
\label{eq:supp-hybrid-mass}
\end{equation}
Consequently, hybrid mass is zero for $r=1$ and, for fixed $\epsilon$,
strictly increases with conflict width $r$.

\paragraph{Proof.}
Assumption F.2 independently selects $A$ with probability $1-\epsilon$ and
$D$ with probability $\epsilon$ at each of the $r$ source-dependent
positions.  The all-$A$ and all-$D$ events therefore have probabilities
$(1-\epsilon)^r$ and $\epsilon^r$.  Every other one of the $2^r-2$ label
vectors belongs to $\calH_R$, which proves Eq.~\eqref{eq:supp-hybrid-mass}.
Let
$\Phi_r(\epsilon)=1-(1-\epsilon)^r-\epsilon^r$.  Holding $\epsilon$ fixed,
\begin{equation}
\Phi_{r+1}(\epsilon)-\Phi_r(\epsilon)
=\epsilon(1-\epsilon)^r+(1-\epsilon)\epsilon^r>0.
\label{eq:supp-width-growth}
\end{equation}
This also gives $\Phi_1(\epsilon)=0$ and proves the stated dose effect.

\paragraph{Partial simultaneous updates.}
For an update set $B\subseteq\{1,\ldots,M\}$ drawn from an unresolved state
that still satisfies Assumption F.1, let $b=|B\cap R|$ and define the marginal
proposal on the labels updated together as
\begin{equation}
Q_{\parallel,B\cap R}(Z_{B\cap R}\mid E)
=\prod_{m\in B\cap R}P_R(Z_m\mid E).
\end{equation}
For $b\geq1$, let $\calH_{B\cap R}$ contain the assignments that use both
source labels; for $b=0$, set $\calH_{B\cap R}=\varnothing$.  Then
\begin{equation}
Q_{\parallel,B\cap R}(\calH_{B\cap R}\mid E)
=
\begin{cases}
0, & b=0,\\
1-(1-\epsilon)^b-\epsilon^b, & b\geq1.
\end{cases}
\label{eq:supp-partial-update}
\end{equation}
Equation~\eqref{eq:supp-partial-update} is zero for $b\leq1$ and positive for
$b\geq2$.  It is a local statement about
one unresolved simultaneous proposal.  Across several denoising steps, the
source posterior and the learned conditionals can change after each update,
so Theorem 1 alone does not determine the final decoded-answer distribution.

\subsection*{Dependence Loss Under the Best Factorized Approximation}

Let $h_2(\epsilon)=-(1-\epsilon)\log(1-\epsilon)
-\epsilon\log\epsilon$ be binary entropy, with $0\log0=0$.

\paragraph{Lemma F.1 (Forward-KL optimal product proposal).}
For any product distribution $Q_R(Z_R)=\prod_{m\in R}q_m(Z_m)$ such that
$q_m$ is positive on the support of $P_{R,m}$,
\begin{equation}
\begin{aligned}
\KL(P_R\|Q_R)
&=\KL\!\left(P_R\middle\|\prod_{m\in R}P_{R,m}\right)\\
&\quad+\sum_{m\in R}\KL(P_{R,m}\|q_m),
\end{aligned}
\label{eq:supp-kl-decomposition}
\end{equation}
where $P_{R,m}$ is the $m$th marginal of $P_R$.
Hence the product of exact marginals in Assumption F.2 minimizes forward KL
over all product distributions.  Under Assumption F.1, its irreducible gap is
\begin{equation}
\KL(P_R\|Q_{\parallel,R})=(r-1)h_2(\epsilon).
\label{eq:supp-forward-kl}
\end{equation}

\paragraph{Proof.}
Using $Q_R=\prod_m q_m$ and adding and subtracting
$\sum_m\log P_{R,m}(Z_m)$ inside the expectation under $P_R$ gives
\begin{align}
\KL(P_R\|Q_R)
&=\mathbb{E}_{P_R}\!\left[
\log\frac{P_R(Z_R)}{\prod_mP_{R,m}(Z_m)}\right]\\
&\quad+\sum_{m\in R}\mathbb{E}_{P_{R,m}}\!\left[
\log\frac{P_{R,m}(Z_m)}{q_m(Z_m)}\right]\\
&=\KL\!\left(P_R\middle\|\prod_mP_{R,m}\right)\\
&\quad+\sum_{m\in R}\KL(P_{R,m}\|q_m),
\end{align}
which proves Eq.~\eqref{eq:supp-kl-decomposition}.  Nonnegativity of each
marginal KL term establishes optimality of the exact-marginal product.
If some $q_m$ vanishes on the support of $P_{R,m}$, then
$\KL(P_R\|Q_R)=+\infty$; hence the same conclusion holds for the product
distributions excluded by the finite-KL premise above.  Because
$0<\epsilon<1$, all marginal probabilities are positive and equality is
possible only when every $q_m=P_{R,m}$, so the minimizer is unique.
Under Assumption F.1, $P_R$ has probabilities $1-\epsilon$ and $\epsilon$
on $A^r$ and $D^r$, whereas $Q_{\parallel,R}$ assigns those vectors
$(1-\epsilon)^r$ and $\epsilon^r$.  Therefore
\begin{align}
\KL(P_R\|Q_{\parallel,R})
&=(1-\epsilon)\log\frac{1-\epsilon}{(1-\epsilon)^r}
+\epsilon\log\frac{\epsilon}{\epsilon^r}\\
&=(r-1)h_2(\epsilon).
\end{align}
Equivalently, this gap is the total correlation
$\sum_{m\in R}H(Z_m)-H(Z_R)=rh_2(\epsilon)-h_2(\epsilon)$.

\subsection*{Corollary 2: Excess Risk from Lost Source Coupling}

\noindent\textbf{Corollary 2.}
Under Assumptions F.1 and F.2, let $A^r$ be the unique accepted source-label
assignment, representing answer $a$ within this restricted abstraction, and
define its sampling risk as $\mathcal{R}(Q)=1-Q(A^r)$.
Let $Q_{\mathrm{causal},R}$ be the exact chain factorization of $P_R$.  Then
$\mathcal{R}_{\mathrm{causal}}=\epsilon$ and, for every $r\geq2$,
\begin{equation}
\mathcal{R}_{\parallel}-\mathcal{R}_{\mathrm{causal}}
=(1-\epsilon)-(1-\epsilon)^r>0.
\label{eq:supp-risk-gap}
\end{equation}

\paragraph{Proof.}
The exact causal chain
$Q_{\mathrm{causal},R}(Z_R)=\prod_jP_R(Z_{m_j}\mid Z_{m_{<j}})$ equals
the joint law $P_R$ for any ordering $(m_1,\ldots,m_r)$.  Its first sampled
label selects $A$ or $D$ with probabilities $1-\epsilon$ and $\epsilon$;
under Eq.~\eqref{eq:supp-source-mixture}, that label identifies the latent
source and every subsequent conditional follows the same branch with
probability one.  The causal chain therefore assigns mass only to $A^r$ and
$D^r$, and its exact-match risk is $\epsilon$.  By Theorem 1, the parallel
proposal is correct with probability $(1-\epsilon)^r$, so its risk is
$1-(1-\epsilon)^r$.  Subtraction proves Eq.~\eqref{eq:supp-risk-gap}.

\paragraph{Weak-conflict regime.}
For fixed $r$ and $\epsilon\to0$,
\begin{equation}
\mathcal{R}_{\parallel}
=1-(1-\epsilon)^r
=r\epsilon+O(\!\epsilon^2),
\qquad
\mathcal{R}_{\mathrm{causal}}=\epsilon.
\label{eq:supp-weak-conflict}
\end{equation}
Thus the leading parallel risk is $r$ times the causal risk, while the excess
risk is $(r-1)\epsilon+O(\epsilon^2)$.  This statement concerns sampling from
the idealized restricted proposals; final greedy or iterative answer accuracy
requires empirical validation.

\subsection*{Heterogeneous Position-Wise Uncertainty}

The equal-$\epsilon$ model provides closed forms, but the combinatorial hybrid
mechanism does not require equal source preference at every position.

\paragraph{Proposition F.2 (Heterogeneous factorized proposal).}
Suppose a factorized proposal selects the correct-source label at position
$j$ with probability $p_j\in(0,1)$.  Over $r\geq1$ source-dependent positions,
its hybrid mass is
\begin{equation}
\Phi(p_1,\ldots,p_r)
=1-\prod_{j=1}^{r}p_j-\prod_{j=1}^{r}(1-p_j).
\label{eq:supp-heterogeneous}
\end{equation}
Adding another unresolved position with $p_{r+1}\in(0,1)$ strictly increases
this mass while the existing $p_1,\ldots,p_r$ are held fixed.

\paragraph{Proof.}
Only the all-$A$ and all-$D$ assignments are source-coherent, with masses
$\prod_jp_j$ and $\prod_j(1-p_j)$, proving
Eq.~\eqref{eq:supp-heterogeneous}.  Let
$A_r=\prod_{j=1}^{r}p_j$ and $D_r=\prod_{j=1}^{r}(1-p_j)$.  Then
\begin{equation}
\begin{aligned}
&\Phi(p_1,\ldots,p_{r+1})-\Phi(p_1,\ldots,p_r)\\
&\qquad=A_r(1-p_{r+1})+D_rp_{r+1}>0.
\end{aligned}
\end{equation}
When the $p_j$ differ, this proposition is a proposal-level robustness
extension; such marginals need not arise from the single shared Bernoulli
source in Assumption F.1.

\subsection*{Matched-Context Confidence--Risk Alignment}

At answer slot $m$, define the method's first-step statistic
\begin{equation}
h_m(q,E)=-\sum_y p_\theta^{(m)}(y\mid q,E)
\log p_\theta^{(m)}(y\mid q,E).
\end{equation}
The block statistic is
\begin{equation}
H(q,E)=\sum_{m=1}^{M}\alpha_m h_m(q,E),
\end{equation}
where the positive weights sum to one.  For the matched real and blank
contexts $E_i^{\mathrm{real}}=\{c_1,c_i\}$ and
$E_i^{\mathrm{blank}}=\{c_1,\blank(c_i)\}$, define
\begin{equation}
\gain_i=H(q,E_i^{\mathrm{blank}})-H(q,E_i^{\mathrm{real}}).
\end{equation}
Geometry matching controls the image-slot layout, but by itself does not imply
that all non-source uncertainty is identical.  The following assumption states
the additional local conditions required for exact alignment.

For a context whose restricted process satisfies Assumption F.1, define its
weighted restricted-label entropy and its residual relative to the actual
full-vocabulary block entropy as
\begin{equation}
\begin{aligned}
H_R^{\mathrm{src}}(E)
&:=\sum_{m\in R}\alpha_m H_{P_{R,m}}(Z_m)\\
&=A_Rh_2(\epsilon_E),\\
C_E&:=H(q,E)-H_R^{\mathrm{src}}(E),\\
A_R&:=\sum_{m\in R}\alpha_m>0.
\end{aligned}
\label{eq:supp-source-residual}
\end{equation}
The quantity $C_E$ is a defined residual; no additive decomposition of the
full-vocabulary entropy is claimed without an additional condition relating
the two matched contexts.

\paragraph{Assumption F.3 (Matched local source basin).}
The real and matched-blank contexts share the same source-token pairs, conflict
set $R$, and weights $\alpha_m$.  In each context, the restricted source
process and its exact-marginal proposal satisfy Assumptions F.1 and F.2 with
context-specific mass $\epsilon_E$, including the continuous degenerate
extension at $\epsilon_E=0$.  Their block entropies decompose as
\begin{equation}
H(q,E)=C+A_Rh_2(\epsilon_E),
\label{eq:supp-entropy-decomposition}
\end{equation}
because the residuals defined in Eq.~\eqref{eq:supp-source-residual} satisfy
$C_{E_i^{\mathrm{real}}}=C_{E_i^{\mathrm{blank}}}=:C$.  Here
$\epsilon_i:=\epsilon_{E_i^{\mathrm{real}}}$ and
$\epsilon_{\blank}:=\epsilon_{E_i^{\mathrm{blank}}}$ both belong to
$[0,\tfrac12]$.  Thus the correct source is
weakly dominant and, within this local scalar decomposition, the block-entropy
difference depends only on the context-specific competing-source mass.
Equality of the two residuals is an explicit local condition, not a consequence
of Assumption F.1 or blank geometry.

\subsection*{Proposition 3: Local Confidence--Risk Alignment}

\noindent\textbf{Proposition 3.}
Under Assumption F.3, for $r\geq2$, positive candidate gain is equivalent to
reducing competing-source mass, first-step hybrid mass, and first-step
factorized exact-match sampling risk:
\begin{equation}
\begin{aligned}
\gain_i>0
&\Longleftrightarrow \epsilon_i<\epsilon_{\blank}\\
&\Longleftrightarrow \Phi_r(\epsilon_i)<\Phi_r(\epsilon_{\blank})\\
&\Longleftrightarrow
\mathcal{R}_{\parallel}(\epsilon_i)
<\mathcal{R}_{\parallel}(\epsilon_{\blank}),
\end{aligned}
\label{eq:supp-confidence-risk}
\end{equation}
where $\Phi_r(\epsilon)=1-(1-\epsilon)^r-\epsilon^r$ and
$\mathcal{R}_{\parallel}(\epsilon)=1-(1-\epsilon)^r$.

\paragraph{Proof.}
By Eq.~\eqref{eq:supp-entropy-decomposition},
\begin{equation}
\gain_i=A_R\left[h_2(\epsilon_{\blank})-h_2(\epsilon_i)\right].
\end{equation}
Because $A_R>0$ and $h_2$ is strictly increasing on
$[0,\tfrac12]$, $\gain_i>0$ if and only if
$\epsilon_i<\epsilon_{\blank}$.  For $r\geq2$,
\begin{equation}
\begin{aligned}
\frac{d}{d\epsilon}\Phi_r(\epsilon)
&=r\left[(1-\epsilon)^{r-1}-\epsilon^{r-1}\right],\\
\frac{d}{d\epsilon}\mathcal{R}_{\parallel}(\epsilon)
&=r(1-\epsilon)^{r-1}>0.
\end{aligned}
\end{equation}
The first derivative is positive on $[0,\tfrac12)$ and vanishes only at the
right endpoint, so $\Phi_r$ is strictly increasing on the full interval
$[0,\tfrac12]$.  The stated equivalences follow.

\paragraph{Approximate matched-context guarantee.}
Retain F.3's shared source-token pairs, conflict set, weights, restricted
process, and correct-source basin, but relax exact residual equality.  Suppose
for some $\delta\geq0$ that
\begin{equation}
H(q,E)=C_E+A_Rh_2(\epsilon_E),
\qquad
|C_{\blank}-C_i|\leq\delta.
\label{eq:supp-approximate-decomposition}
\end{equation}
where $C_i:=C_{E_i^{\mathrm{real}}}$ and
$C_{\blank}:=C_{E_i^{\mathrm{blank}}}$.
Then
\begin{equation}
\left|\gain_i-A_R\left[
h_2(\epsilon_{\blank})-h_2(\epsilon_i)\right]\right|\leq\delta.
\label{eq:supp-approximate-gain}
\end{equation}
Consequently, $\gain_i>\delta$ is sufficient for
$\epsilon_i<\epsilon_{\blank}$ and hence for lower first-step hybrid mass and
sampling risk.  Conversely, a reduction in competing-source mass guarantees
$\gain_i>0$ whenever its ideal entropy decrease
$A_R[h_2(\epsilon_{\blank})-h_2(\epsilon_i)]$ exceeds $\delta$.
The implemented zero threshold is therefore exactly certified by
Proposition 3 only within the matched basin; outside it, the gain remains an
empirically validated surrogate rather than a universal correctness
certificate.

\begin{table*}[!t]
\centering
\small
\setlength{\tabcolsep}{3.5pt}
\begin{tabular}{llccccc}
\toprule
Model & Method & ChartQA & InfoChartQA & DocVQA & InfoVQA & TATDQA \\
\midrule
\multirow{7}{*}{LLaDA2.0-Uni}
& Fixed top-2 & 24.60 & 18.96 & 30.88 & 30.12 & 11.52 \\
& Frozen Semantic Top-2 & 25.88 & 17.16 & 31.64 & 32.80 & 12.16 \\
& SPREAD Top-2 & 26.12 & 18.68 & 33.56 & 32.72 & 12.60 \\
& SARDI Top-2 & 27.20 & 19.04 & 36.12 & 34.88 & 12.88 \\
& UOVES Top-2 & 26.84 & 20.00 & \textbf{37.16} & 35.92 & 13.76 \\
& Answer-UQ Top-1 & \underline{30.72} & \underline{20.68} & 32.92 & \underline{36.08} & \underline{14.08} \\
& \method{} & \textbf{31.12} & \textbf{21.16} & \underline{36.72} & \textbf{42.60} & \textbf{18.96} \\
\midrule
\multirow{7}{*}{LLaDA-V}
& Fixed top-2 & 13.32 & 12.60 & 15.88 & 18.32 & 8.72 \\
& Frozen Semantic Top-2 & 13.68 & 13.20 & 16.40 & 18.52 & 8.72 \\
& SPREAD Top-2 & 13.84 & 12.24 & 16.00 & 16.84 & 8.88 \\
& SARDI Top-2 & 12.88 & 11.08 & 15.56 & 17.92 & 8.00 \\
& UOVES Top-2 & 12.60 & 11.88 & 14.84 & 17.92 & 8.92 \\
& Answer-UQ Top-1 & \underline{20.36} & \underline{15.88} & \underline{21.04} & \underline{26.16} & \underline{10.56} \\
& \method{} & \textbf{20.48} & \textbf{16.28} & \textbf{24.44} & \textbf{30.84} & \textbf{14.04} \\
\midrule
\multirow{7}{*}{Dream-VL}
& Fixed top-2 & 33.60 & \textbf{25.32} & \underline{38.48} & 47.64 & 17.00 \\
& Frozen Semantic Top-2 & 32.52 & 22.60 & 34.96 & 44.68 & 16.60 \\
& SPREAD Top-2 & 32.68 & 24.24 & 36.72 & 46.04 & 17.00 \\
& SARDI Top-2 & \underline{34.40} & 24.60 & \textbf{39.32} & 49.28 & 16.80 \\
& UOVES Top-2 & \textbf{34.60} & \textbf{25.32} & \textbf{39.32} & \underline{49.32} & \underline{18.16} \\
& Answer-UQ Top-1 & 32.52 & 24.56 & 37.16 & 48.36 & 17.32 \\
& \method{} & \underline{34.40} & \underline{25.08} & 37.60 & \textbf{49.68} & \textbf{21.16} \\
\bottomrule
\end{tabular}
\caption{Answer accuracy (\%) across three DLM backends. Within each backend and dataset, bold and underlined denote the best and second-best values; ties share formatting. All methods use a candidate pool of $k=3$, except fixed top-2.}
\label{tab:training-free-all-backends}
\end{table*}

\section{Appendix G: Controlled-Study Details}
\label{app:controlled-details}

\paragraph{Synthetic source-conflict validation.}
The confirmatory study uses 32 independent base templates.
Each correct page describes the target unit and supports answer $a$, while a layout-matched hard negative describes another unit and supports $d$.
Conflict slots are added in a nested seeded order for $r\in\{1,2,4\}$, and both correct--distractor and distractor--correct page orders are retained.
At each conflicting position, probability is renormalized over the correct-source and conflicting-source tokens; the reported quantity is therefore restricted source mass rather than full-vocabulary sequence probability.
Inference uses 10,000 base-template bootstrap resamples, treating conflict width, order, and model construction as repeated conditions within each template.

All 576 expected rows ($32$ base templates $\times$ $3$ conflict widths $\times$ $2$ page orders $\times$ $3$ model constructions) were present with unique keys, no inference error occurred, the maximum mass-sum error was $1.11\times10^{-16}$, and the largest absolute $r=1$ hybrid mass was $5.55\times10^{-17}$.
The parallel-minus-sequential effect was positive for 31 of 32 templates with paired $d_z=1.62$.

\paragraph{Free-generation validation of cross-source hybrids.}
Both models answered all 36 single-page conditions coherently for each source.
Qwen produced a coherent correct answer in all 72 two-page conditions.
Native greedy LLaDA produced a literal cross-source hybrid in 11 of 48 conditions with $r\geq2$.
First-step restricted hybrid mass was higher for realized hybrids than for other outputs (0.735 vs. 0.393) and yielded an exploratory AUROC of 0.801.
A block-size-one LLaDA schedule also produced 11 hybrids, indicating that changing commit order alone did not remove source conflict already present in the masked conditionals.

\paragraph{Natural-page order breakdown.}
Table~\ref{tab:controlled-order} reports the order-specific values underlying the semantic-conflict averages in the main paper.

\begin{table}[!ht]
\centering
\small
\setlength{\tabcolsep}{1.2pt}
\begin{tabular}{lcccc}
\toprule
Condition & LLaDA2.0 & LLaDA-V & Dream-VL & Qwen AR \\
\midrule
Correct only & 75.88 & 46.16 & 79.24 & 77.88 \\
Correct + wrong & 44.92 & 16.68 & 62.68 & 61.08 \\
Wrong + correct & 60.72 & 33.52 & 66.04 & 63.04 \\
\bottomrule
\end{tabular}
\caption{Order-specific controlled answer accuracy (\%) on ChartQA across three DLMs and an autoregressive baseline.}
\label{tab:controlled-order}
\end{table}
\FloatBarrier

\section{Appendix H: Training-Free Baseline Details}
\label{app:training-free}

\paragraph{UOVES.}
Qwen3-VL-2B-Instruct \citep{bai2025qwen3vl} judges each of the three candidates independently with a binary helpfulness prompt.
The raw first-token \texttt{True} logit ranks candidates, and the top two are passed to the target DLM \citep{luo2026uoves}.

\paragraph{Answer-UQ.}
The target DLM generates once from each candidate separately.
For each committed answer position, we compute full-vocabulary entropy at its commit step, average across valid answer positions, and return the answer with minimum mean entropy.

\paragraph{Frozen Semantic.}
SigLIP2 SO400M \citep{tschannen2025siglip2} independently encodes the question and each candidate.
The two candidates with the highest normalized cosine similarity are passed to the generator.

\paragraph{SPREAD.}
The VisRAG top two remain fixed.
At each denoising update, answer-position hidden states are ranked by cosine relevance to a pooled query representation, and the native reveal budget commits the most relevant positions first \citep{yu2026spread}.

\paragraph{SARDI.}
Generation starts from the VisRAG top two.
At each update, tentative answer tokens are appended to the question, VisRAG retrieves a new top-two context, and committed answer tokens are transferred into the rebuilt multimodal state \citep{junger2026sardi}.

UOVES, Answer-UQ, and Frozen Semantic follow comparator logic released with UOVES; SPREAD and SARDI are visual adaptations of text DLM-RAG methods.
All methods use the same eight-token answer region and unified scorer as \method{}.

\section{Appendix I: Complete Training-Free Baseline Comparison}
\label{app:training-free-results}

Table~\ref{tab:training-free-all-backends} extends the main paper's LLaDA2.0-Uni comparison to all three DLM backends under the same evaluation protocol.

Across the complete comparison, \method{} achieves the best accuracy in 11 of the 15 model--dataset settings, including all five settings with LLaDA-V. It also remains strongest on InfoVQA and TATDQA with Dream-VL, while UOVES performs best or ties for best on the other three Dream-VL settings.

\section{Appendix J: Efficiency}
\label{app:efficiency}

Table~\ref{tab:latency} reports end-to-end latency measured on a single NVIDIA H200 GPU. \method{} adds first-step probe forward passes before final generation.

\begin{table}[!ht]
\centering
\small
\setlength{\tabcolsep}{3pt}
\begin{tabular}{lcccc}
\toprule
Backend & top-1 & top-2 & top-3 & \method{} \\
\midrule
LLaDA2.0-Uni & 2.35 & 3.17 & 3.73 & 3.63 \\
LLaDA-V & 4.70 & 8.35 & 16.86 & 8.01 \\
Dream-VL & 3.72 & 12.16 & 10.80 & 7.59 \\
\bottomrule
\end{tabular}
\caption{End-to-end inference latency in seconds per question. \method{} uses $k=3$.}
\label{tab:latency}
\end{table}

For candidate-pool size $k$, the worst case computes $H(q,c_1)$, real and control entropies for $c_2,\ldots,c_k$, and one final generation pass; probe cost therefore grows linearly with $k$.
The implementation caches visual embeddings when the backend exposes reusable states, but different DLMs provide different cache interfaces.

Relative to fixed multi-image inference, \method{} is close to top-2 latency on LLaDA2.0-Uni (3.63 vs. 3.17 seconds) and is faster than top-2 on LLaDA-V and Dream-VL (8.01 vs. 8.35 and 7.59 vs. 12.16 seconds).
It is also faster than fixed top-3 inference on all three backends.
Together with the main accuracy results, these measurements show that \method{} makes more effective use of the retrieved candidate pool: it converts additional candidate availability into higher answer accuracy while keeping inference cost comparable to or below fixed multi-image input.
\FloatBarrier

\bibliography{references}